\documentclass[conference]{IEEEtran}
\IEEEoverridecommandlockouts

\usepackage{cite}
\usepackage{amsmath,amssymb,amsfonts}
\usepackage{algorithmic}
\usepackage{algorithm}
\usepackage{graphicx}
\usepackage{textcomp}
\usepackage{xcolor}
\usepackage{hyperref}
\usepackage{booktabs}
\usepackage{multirow}
\usepackage{url}
\usepackage{balance}
\usepackage{stfloats}

\def\BibTeX{{\rm B\kern-.05em{\sc i\kern-.025em b}\kern-.08em
    T\kern-.1667em\lower.7ex\hbox{E}\kern-.125emX}}

\begin{document}

\title{MR-SLAM: Immersive Spatial Supervision for Multi-Robot Mapping via Mixed Reality
\thanks{This work was supported by Horizon 2020 (EU Commission) project InnoGuard, Marie Sk\l{}odowska-Curie Actions Doctoral Networks (HORIZON-MSCA-2023-DN), the SNSF SwarmOps project (No.~200021\_219732), and the Hasler Foundation project on UAV Reliability and Societal Trust (No.~2025-02-27-311).}}

\author{
\IEEEauthorblockN{Prakash Aryan\IEEEauthorrefmark{1}, Cem Erdogdu\IEEEauthorrefmark{1}, Kavinaya Kumarchokkappan\IEEEauthorrefmark{1}, Timo Kehrer\IEEEauthorrefmark{1}, Sebastiano Panichella\IEEEauthorrefmark{1}\IEEEauthorrefmark{2}}
\IEEEauthorblockA{\IEEEauthorrefmark{1}University of Bern, Bern, Switzerland}
\IEEEauthorblockA{\IEEEauthorrefmark{2}AI4I -- The Italian Institute of Artificial Intelligence, Turin, Italy}
\IEEEauthorblockA{prakash.aryan@unibe.ch, \{cem.erdogdu, kavinaya.kumarchokkappan\}@students.unibe.ch,}
\IEEEauthorblockA{timo.kehrer@unibe.ch, \{sebastiano.panichella@unibe.ch, sebastiano.panichella@ai4i.it\}}
}

\maketitle

\begin{abstract}
Operating a multi-robot fleet for simultaneous localization and mapping (SLAM) in applications such as building inspection or warehouse-aisle monitoring requires the operator to maintain spatial awareness of each robot's position and mapping state, a task that scales poorly on conventional 2D interfaces. We present MR-SLAM, a mixed reality (MR) system in which an operator wearing a Meta Quest~3 headset teleoperates three simulated TurtleBot3 robots through a passthrough view with real-world occlusion, while spatially anchored dashboard panels report mapping progress in situ. Each robot runs an independent SLAM Toolbox instance whose occupancy grid is merged in real time on a Robot Operating System~2 (ROS~2) back-end. Across five 9-minute evaluation sessions, the system delivered scans at $8.83 \pm 0.16$\,Hz, mapped $17.9 \pm 0.8$\,m$^2$ of merged occupancy, and reached $94.7 \pm 0.5\%$ cross-instance occupancy consistency across robot pairs. An additional session recorded 6.3\,ms median transform jitter and 26.7\,m$^2$ coverage of a 41\,m$^2$ grid. We position MR-SLAM as a reference implementation for combining passthrough mixed reality supervision with multi-robot SLAM on consumer hardware.
\end{abstract}


\section{Introduction}
\label{sec:intro}

Simultaneous localization and mapping (SLAM) allows mobile robots to operate without external positioning infrastructure~\cite{cadena_past_2016, durrant-whyte_simultaneous_2006}. The problem was first formulated as a joint state estimation task~\cite{smith_estimating_2013} and later grounded in probabilistic robotics~\cite{thrun_probabilistic_2005}. Extending SLAM to multi-robot teams enables faster area coverage and fault tolerance, but introduces challenges in communication, map fusion, and operator supervision~\cite{lajoie_towards_2022}. These challenges are seen in time-critical applications such as building inspection and warehouse inventory, where an on-site operator must coordinate multiple robots from a single station. The SLAM back-end has matured from graph-based optimizers~\cite{grisetti_tutorial_2010} to distributed systems~\cite{lajoie_door-slam_2020, lajoie_swarm-slam_2024}, yet the operator interface for multi-robot supervision during mapping has received limited attention.

Robot teleoperation typically relies on desktop workstations with 2D displays and joystick or keyboard input~\cite{sheridan_telerobotics_1992,surrealist}. These interfaces require the operator to mentally reconstruct 3D spatial relationships from flat views, and this cognitive burden grows with the number of robots~\cite{wonsick_systematic_2020}. Virtual Reality (VR), augmented reality (AR), and mixed reality (MR) head-mounted displays (HMDs) have been shown to improve situational awareness in single-robot teleoperation~\cite{suzuki_augmented_2022, walker_virtual_2022, hedayati_improving_2018} as well as facilitate safety assessment of autonomous systems \cite{VR-study}. Walker et al.~\cite{walker_cyber-physical_2024} reported a 28\% improvement in operator effectiveness using an MR interface for single-robot navigation. However, none addresses the coordination challenges of supervising multiple robots during active SLAM.

In prior work, we developed SimNav-XR~\cite{aryan_simnav-xr_2026}, an extended reality platform for single-robot simulation integrating Unity, ROS~2, and the Meta Quest~3. SimNav-XR demonstrated MR-based robot control with LiDAR simulation and Nav2 integration but was limited to a single robot without real-time SLAM.

This paper presents MR-SLAM, which extends SimNav-XR to multi-robot SLAM with three contributions: (1)~a passthrough MR interface on the Meta Quest~3 rendering three simulated TurtleBot3 robots with raycasted LiDAR, spatial occlusion, and joystick teleoperation; (2)~a multi-robot SLAM pipeline with independent SLAM Toolbox~\cite{macenski_slam_2021} instances and real-time map merging via \texttt{multirobot\_map\_merge}~\cite{birk_merging_2006, carpin_fast_2008}; and (3)~a spatially anchored dashboard design displaying coverage, overlap, scan rate, and latency metrics. The system targets scenarios where an operator supervises a small robot fleet from a fixed location, such as inspecting a building wing or monitoring warehouse aisles.

\section{Related Work}
\label{sec:related}

\subsection{Multi-Robot SLAM}

Multi-robot SLAM architectures range from centralized to fully decentralized~\cite{lajoie_towards_2022,PrakashPanichella2026}. Centralized systems such as CCM-SLAM~\cite{schmuck_ccm-slam_2019} aggregate data at a server for joint optimization; this simplifies map fusion but introduces a communication bottleneck and a single point of failure. Distributed systems such as DOOR-SLAM~\cite{lajoie_door-slam_2020} and Kimera-Multi~\cite{chang_kimera-multi_2021, tian_kimera-multi_2021} perform peer-to-peer pose-graph optimization with outlier rejection, improving scalability at the cost of increased synchronization complexity. Swarm-SLAM~\cite{lajoie_swarm-slam_2024} provided a ROS~2 decentralized framework supporting multiple sensor modalities, and LAMP~\cite{ebadi_lamp_2020} and LAMP~2.0~\cite{chang_lamp_2022} addressed subterranean multi-robot SLAM during the DARPA SubT Challenge.

Occupancy grid mapping~\cite{elfes_using_1989, moravec_high_1985} remains standard for 2D LiDAR. ORB-SLAM2~\cite{mur-artal_orb-slam2_2017} established feature-based visual SLAM, and Google Cartographer~\cite{hess_real-time_2016} demonstrated real-time 2D LiDAR SLAM with branch-and-bound loop closure. For ROS~2 specifically, Macenski et al.\ developed SLAM Toolbox~\cite{macenski_slam_2021} as a lifelong 2D SLAM solution, and benchmarked modern visual SLAM approaches for Nav2 integration~\cite{merzlyakov_comparison_2021}. Map merging for occupancy grids was formalized by Carpin and Birk~\cite{carpin_map_2005, birk_merging_2006} using spectral methods and stochastic search, and extended with Hough-transform acceleration~\cite{carpin_fast_2008}. The \texttt{multirobot\_map\_merge} ROS package has been applied in multi-TurtleBot3 deployments~\cite{nica_development_2022, vascak_map_2023}.

\subsection{Mixed Reality for Robot Teleoperation}

Surveys by Suzuki et al.~\cite{suzuki_augmented_2022} and Walker et al.~\cite{walker_virtual_2022} catalogued AR/MR strategies for human-robot interaction (HRI). Early work demonstrated intent communication via HMD overlays~\cite{rosen_communicating_2017, hedayati_improving_2018}. Whitney et al.~\cite{whitney_ros_2018} introduced ROS Reality for Unity-based virtual reality (VR) teleoperation. On the Meta Quest platform, Quest2ROS~\cite{welle_quest2ros_2024} measured 0.46\,mm tracking accuracy with 82\,ms latency, and Quest2ROS2~\cite{li_quest2ros2_2026} extended this to ROS~2. Morando and Loianno~\cite{morando_spatial_2024} demonstrated MR-based human-drone collaboration with spatial awareness sharing. Rold\'{a}n et al.~\cite{roldan_multi-robot_2019} addressed multi-robot VR monitoring without SLAM, and the Cyber-Physical Control Room~\cite{walker_cyber-physical_2024} provided MR perspectives for single-robot teleoperation. However, none addresses the cognitive and coordination challenges of supervising multiple robots during active SLAM.

\subsection{ROS--Unity Integration}

The Unity Robotics Hub~\cite{noauthor_unity-technologiesunity-robotics-hub_2026} provides \texttt{ros\_tcp\_connector} for Unity--ROS communication. Allspaw et al.~\cite{allspaw_comparing_2023} benchmarked Unity--ROS bridges and found \texttt{ros\_tcp\_connector} offers the best latency-integration trade-off. Platt and Ricks~\cite{platt_comparative_2022} showed that Unity matches or exceeds Gazebo for SLAM simulation with better scalability. These integration frameworks enable real-time bidirectional communication, yet none has been applied to multi-robot SLAM with an immersive operator interface.

\subsection{Positioning of This Work}

Table~\ref{tab:comparison} compares MR-SLAM with related systems. Prior work addresses either MR-based single-robot teleoperation~\cite{walker_cyber-physical_2024, morando_spatial_2024, welle_quest2ros_2024} or multi-robot SLAM without an MR interface~\cite{lajoie_swarm-slam_2024, nica_development_2022}. Rold\'{a}n et al.~\cite{roldan_multi-robot_2019} built a multi-robot VR interface but did not integrate SLAM. Patel et al.~\cite{patel_towards_2024} explored MR for multi-robot interaction but without a SLAM pipeline. In contrast, MR-SLAM uniquely combines passthrough MR visualization, real-time multi-robot SLAM with occupancy grid merging, and spatial performance feedback into a unified system, enabling operators to supervise mapping tasks directly within their physical environment.

\begin{table}[!htbp]
\centering
\caption{Comparison with Related Systems}
\label{tab:comparison}
\small
\begin{tabular}{@{}lcccc@{}}
\toprule
\textbf{System} & \textbf{MR} & \textbf{Multi} & \textbf{SLAM} & \textbf{Dash.} \\
 & \textbf{Pass.} & \textbf{Robot} & & \\
\midrule
Cyber-Phys.\ CR~\cite{walker_cyber-physical_2024} & \checkmark & -- & -- & \checkmark \\
Quest2ROS~\cite{welle_quest2ros_2024} & -- & -- & -- & -- \\
Morando et al.~\cite{morando_spatial_2024} & \checkmark & -- & -- & -- \\
Rold\'{a}n et al.~\cite{roldan_multi-robot_2019} & -- & \checkmark & -- & \checkmark \\
Patel et al.~\cite{patel_towards_2024} & \checkmark & \checkmark & -- & -- \\
Swarm-SLAM~\cite{lajoie_swarm-slam_2024} & -- & \checkmark & \checkmark & -- \\
Nica et al.~\cite{nica_development_2022} & -- & \checkmark & \checkmark & -- \\
SimNav-XR~\cite{aryan_simnav-xr_2026} & \checkmark & -- & \checkmark & -- \\
\textbf{MR-SLAM (ours)} & \checkmark & \checkmark & \checkmark & \checkmark \\
\bottomrule
\end{tabular}
\end{table}

\section{System Architecture}
\label{sec:system}

\begin{figure*}[!htbp]
\centering
\includegraphics[width=\textwidth]{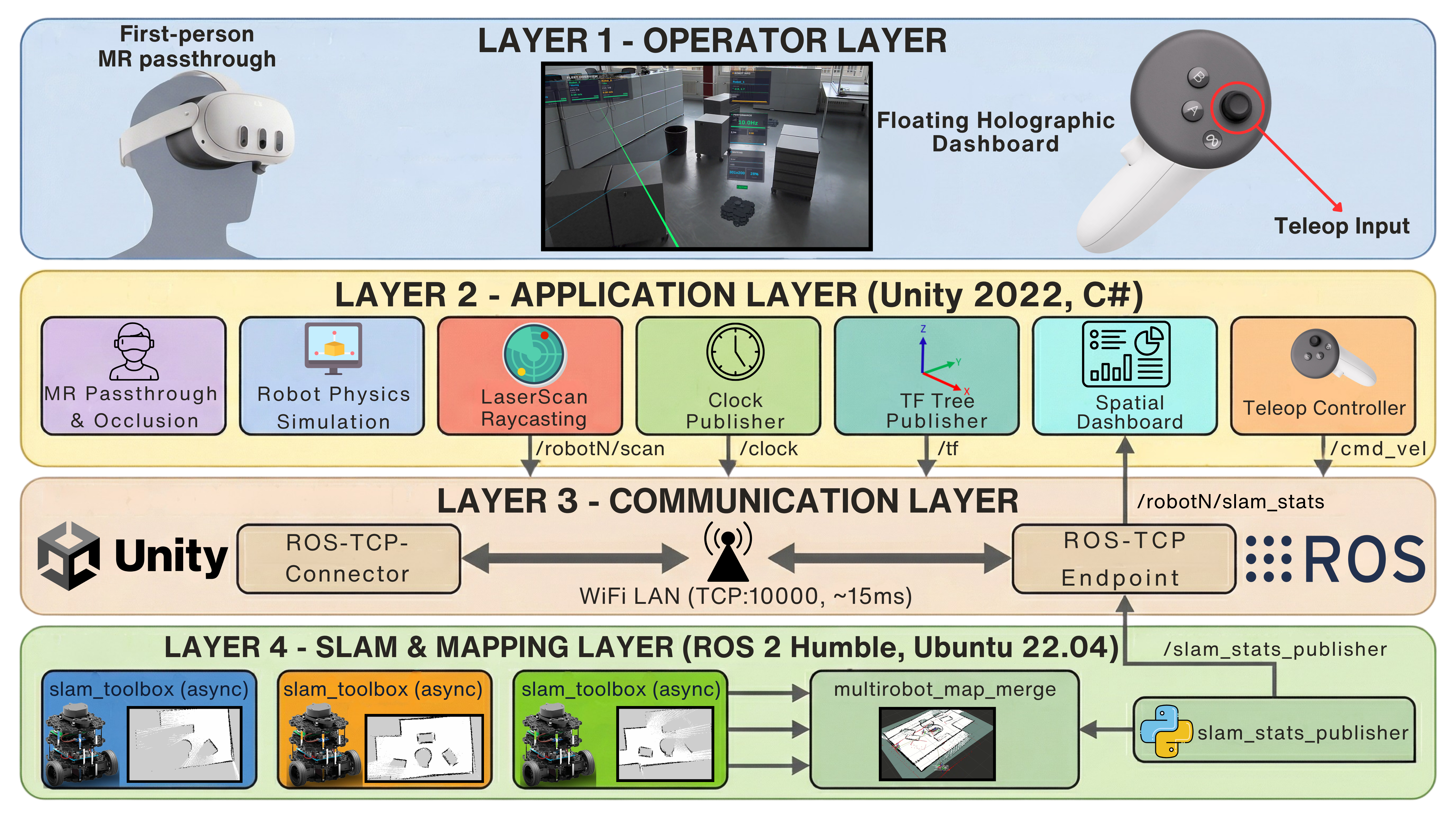}
\caption{MR-SLAM system architecture. Layer~1: operator with Meta Quest~3 passthrough MR. Layer~2: Unity application with robot physics, LiDAR raycasting, TF publishing, spatial dashboard, and teleop controller. Layer~3: \texttt{ros\_tcp\_connector} bridge over WiFi (TCP:10000). Layer~4: three independent SLAM Toolbox instances produce per-robot maps that are fused by \texttt{multirobot\_map\_merge}.}
\label{fig:sysarch}
\end{figure*}

MR-SLAM consists of a Unity-based MR application on the Meta Quest~3 and a ROS~2 SLAM back-end on an Ubuntu~22.04 laptop, connected over WiFi. All robots are fully simulated within Unity; no physical robots are used. The operator wears the headset in a real room, sees virtual robots overlaid on the physical environment through passthrough rendering, and issues velocity commands via the controller. These commands drive simulated differential-drive physics, generating LiDAR scans and transform data streamed to ROS~2 for SLAM processing; resulting maps and statistics are sent back for dashboard display. Fig.~\ref{fig:sysarch} shows the layered architecture.

\subsection{MR Application Layer}

\begin{figure}[!htbp]
\centering
\includegraphics[width=\columnwidth]{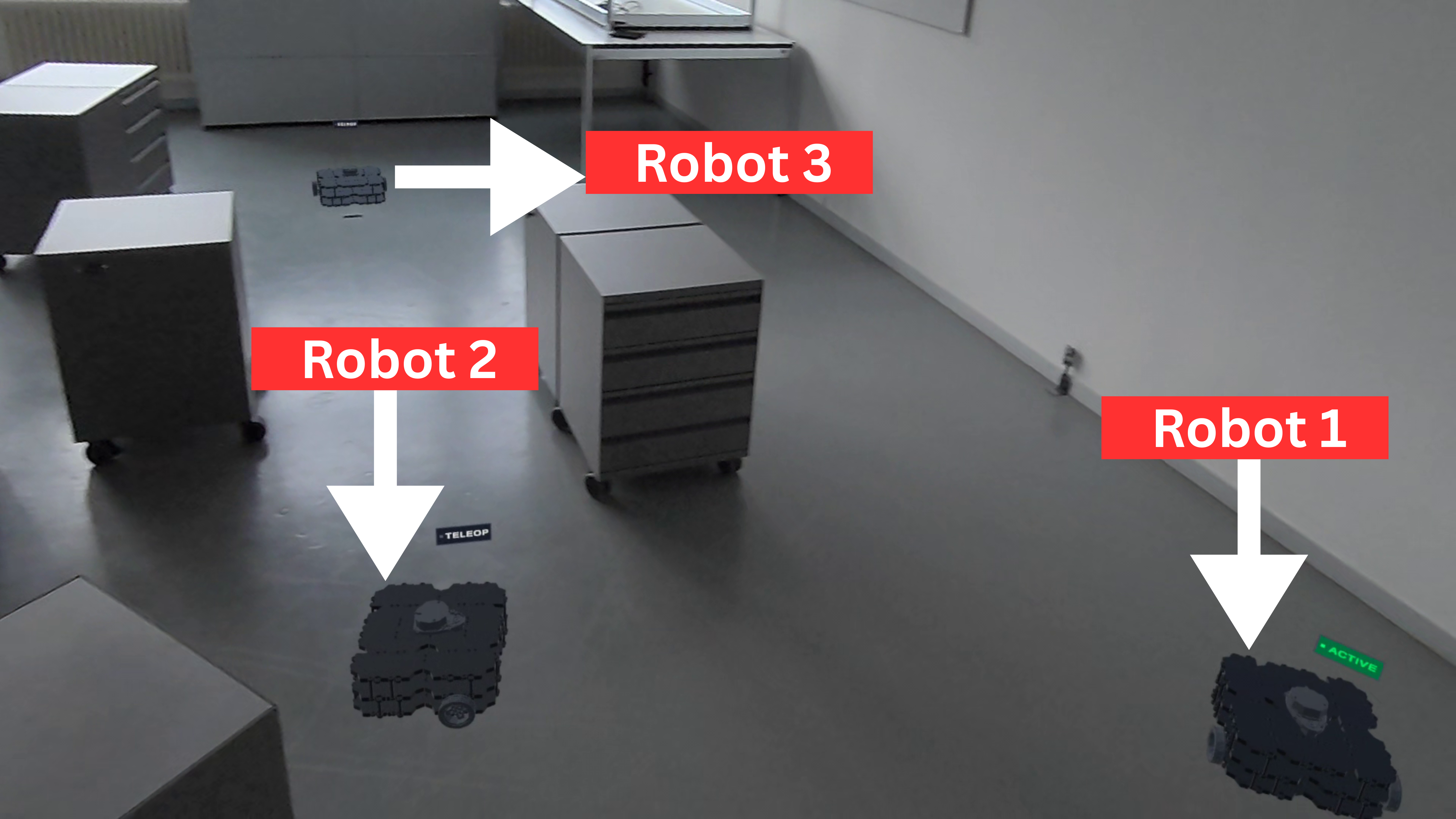}
\caption{Three simulated TurtleBot3 robots placed in the laboratory environment as seen through Meta Quest~3 passthrough. MRUK spatial occlusion hides robots behind real-world furniture.}
\label{fig:three_robots}
\end{figure}

The application runs in Unity~2022~LTS on Android (Meta Quest~3). Meta's Mixed Reality Utility Kit (MRUK) provides room-scale passthrough and spatial occlusion, using on-device AI to reconstruct a spatial mesh of the physical environment from depth and color data: virtual robots behind real furniture are correctly hidden (Fig.~\ref{fig:three_robots}). Three TurtleBot3 Burger models~\cite{noauthor_turtlebot3_nodate, amsters_turtlebot_2020} use differential-drive physics controllers with articulated wheel joints.

Each robot carries a simulated 2D LiDAR implemented via Unity raycasting---180 rays over a 90$^\circ$ forward-facing arc at 10\,Hz---producing \texttt{LaserScan} messages on namespaced topics (\texttt{/robot1/scan}, etc.). This configuration approximates the forward field of view of a planar LiDAR while maintaining real-time performance on the headset's mobile processor. A TF publisher broadcasts \texttt{odom}$\rightarrow$\texttt{base\_footprint} and child frames at 20\,Hz. A clock publisher sends simulation time at 100\,Hz via \texttt{/clock}. Teleoperation uses the Meta Quest~3 right Touch controller thumbstick to issue \texttt{Twist} commands, clamped to 0.15\,m/s linear and 1.5\,rad/s angular velocity.

\subsection{Spatial Dashboard}

\begin{figure}[!htbp]
\centering
\includegraphics[width=\columnwidth]{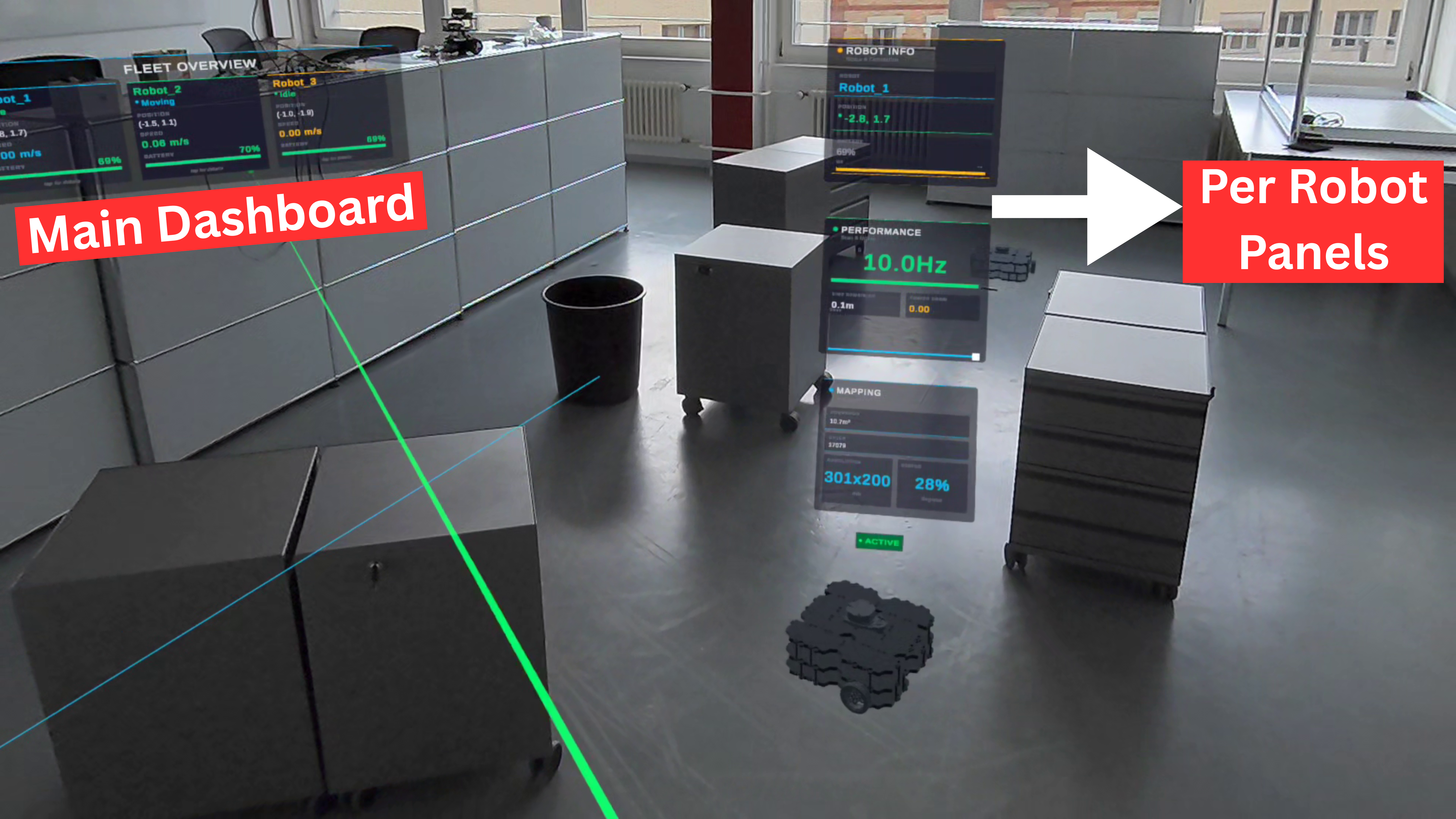}
\caption{In-headset spatial dashboard viewed through passthrough MR. Left: fleet overview. Right: per-robot panels showing scan rate (10.0\,Hz), map dimensions, and coverage.}
\label{fig:dashboard}
\end{figure}

World-space Canvas panels anchored in the MR environment provide real-time fleet telemetry (Fig.~\ref{fig:dashboard}). A fleet overview panel shows each robot's connection status and scan rate. Per-robot panels display map dimensions, coverage area, and the latest scan timestamp, updated at 1\,Hz from the \texttt{slam\_stats\_publisher} node.

\subsection{Communication Bridge}

Bidirectional communication uses \texttt{ros\_tcp\_connector}~\cite{noauthor_unity-technologiesunity-robotics-hub_2026} (Unity side) and \texttt{ros\_tcp\_endpoint} (ROS~2 side) over TCP port~10000. Table~\ref{tab:topics} lists the primary topics.

\begin{table}[!htbp]
\centering
\caption{Primary ROS~2 Topics}
\label{tab:topics}
\small
\begin{tabular}{@{}llcc@{}}
\toprule
\textbf{Topic} & \textbf{Type} & \textbf{Rate} & \textbf{Dir.} \\
\midrule
\texttt{/robotN/scan} & LaserScan & 10\,Hz & U$\rightarrow$R \\
\texttt{/tf} & TFMessage & 20\,Hz & U$\rightarrow$R \\
\texttt{/clock} & Clock & 100\,Hz & U$\rightarrow$R \\
\texttt{/cmd\_vel[\_N]} & Twist & on-demand & U$\rightarrow$R \\
\texttt{/robotN/map} & OccupancyGrid & $\sim$1\,Hz & R$\rightarrow$U \\
\texttt{/robotN/slam\_stats} & Float32MultiArray & 1\,Hz & R$\rightarrow$U \\
\bottomrule
\multicolumn{4}{@{}l}{\footnotesize U = Unity (Meta Quest~3), R = ROS~2. N $\in$ \{1, 2, 3\}.}
\end{tabular}
\end{table}

\subsection{SLAM and Map Merging}

On the ROS~2 back-end (Intel i5, 16\,GB RAM), three SLAM Toolbox~\cite{macenski_slam_2021} instances run in asynchronous mode, each subscribing to a namespaced scan topic. Asynchronous mode was selected for its tolerance of variable-latency scan delivery over WiFi and its support for lifelong mapping with serialization. The \texttt{multirobot\_map\_merge} node subscribes to all three maps and produces a merged occupancy grid. Because the robots share a common \texttt{map} frame, the merge uses known initial poses rather than the more costly unknown-pose spectral matching~\cite{carpin_fast_2008}.

\begin{figure*}[!htbp]
\centering
\includegraphics[width=0.85\textwidth]{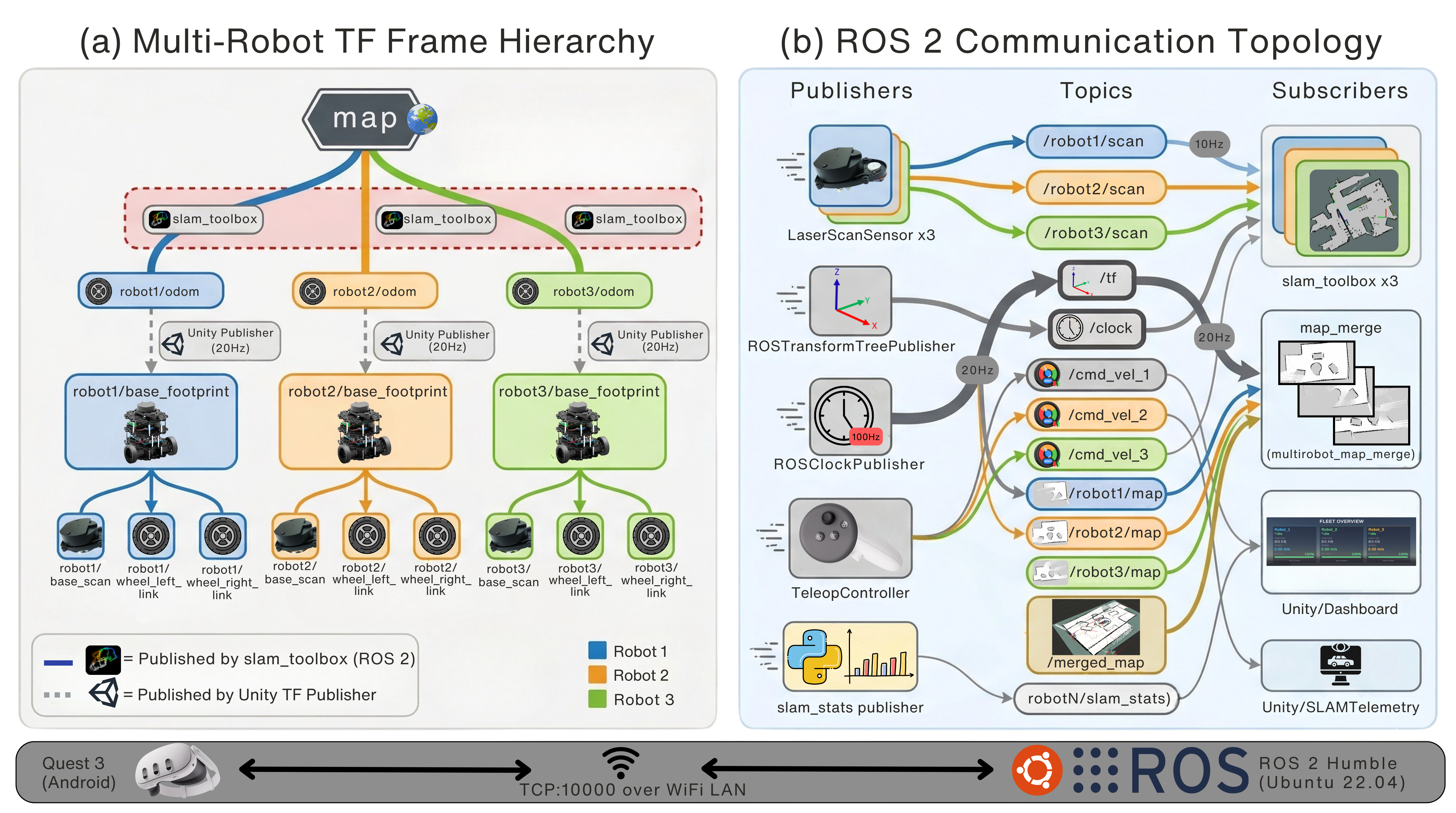}
\caption{(a)~TF frame hierarchy: solid arrows = SLAM-published (\texttt{map}$\rightarrow$\texttt{odom}), dashed = Unity-published. Red callout marks the partitioning constraint. (b)~ROS~2 topic graph with publishers, subscribers, and data rates.}
\label{fig:tf_ros}
\end{figure*}

Algorithm~\ref{alg:tf} details the complete MR-SLAM pipeline from sensor simulation through SLAM to dashboard update. A critical constraint is that Unity publishes only the \texttt{odom}$\rightarrow$\texttt{base\_footprint} subtree while SLAM Toolbox publishes \texttt{map}$\rightarrow$\texttt{odom} (line~5). Allowing both to publish overlapping transforms caused TF tree conflicts and SLAM divergence. Fig.~\ref{fig:tf_ros} illustrates the TF frame hierarchy and ROS~2 communication topology, showing this partitioning and the topic-level data flow.

\begin{algorithm}[!htbp]
\caption{MR-SLAM End-to-End Pipeline}
\label{alg:tf}
\begin{algorithmic}[1]
\REQUIRE $N$ robots in Unity, SLAM Toolbox $\times N$, map\_merge node
\ENSURE Merged occupancy grid, live dashboard metrics
\STATE \textbf{// Sensor and TF Bridge (Unity $\rightarrow$ ROS~2)}
\FOR{each robot $r \in \{1, \ldots, N\}$}
    \STATE Cast 180 rays over 90$^\circ$ arc at 10\,Hz $\rightarrow$ \texttt{/robot$_r$/scan}
    \STATE Publish $T_{r/\text{odom} \rightarrow r/\text{base\_footprint}}$ at 20\,Hz
    \STATE \textbf{Skip} $T_{\text{map} \rightarrow r/\text{odom}}$ \COMMENT{SLAM owns this}
\ENDFOR
\STATE Publish \texttt{/clock} at 100\,Hz with wall-clock offset $\delta_t$
\STATE \textbf{// Per-Robot SLAM (ROS~2)}
\FOR{each robot $r$}
    \STATE SLAM Toolbox$_r$ $\leftarrow$ subscribe \texttt{/robot$_r$/scan}
    \STATE SLAM Toolbox$_r$ $\rightarrow$ publish \texttt{/robot$_r$/map}, $T_{\text{map} \rightarrow r/\text{odom}}$
\ENDFOR
\STATE \textbf{// Map Fusion and Telemetry (ROS~2)}
\STATE map\_merge $\leftarrow$ \{/robot$_1$/map, \ldots, /robot$_N$/map\}
\STATE map\_merge $\rightarrow$ \texttt{/merged\_map} using known initial poses
\FOR{each robot $r$}
    \STATE Classify cells: free ($v{=}0$), occupied ($v{>}50$), unknown ($v{=}{-}1$)
    \STATE Publish $[n_f, n_o, n_u, n_t, w, h, res]$ to \texttt{/robot$_r$/slam\_stats}
\ENDFOR
\STATE \textbf{// Dashboard Update (Unity)}
\FOR{each robot $r$}
    \STATE coverage$_r \leftarrow (n_f + n_o) \times res^2$
    \STATE scan\_rate$_r \leftarrow$ LaserScanSensor publish rate
\ENDFOR
\STATE \textbf{// Operator Teleop (Unity $\leftarrow$ Meta Quest~3 controller)}
\STATE Read joystick $\rightarrow$ apply deadzone $\rightarrow$ clamp $(v, \omega)$
\STATE Publish \texttt{Twist} to selected robot's \texttt{/cmd\_vel}
\end{algorithmic}
\end{algorithm}

A further integration challenge arose from a Quality of Service (QoS) mismatch between \texttt{ros\_tcp\_connector}'s RELIABLE clock and SLAM Toolbox's BEST\_EFFORT subscription, which caused silent clock drops. A 5.0\,s clock offset applied in the Unity \texttt{WallClockTime} component compensates for the accumulated drift between simulation and wall-clock time.

A custom \texttt{slam\_stats\_publisher.py} node queries each robot's map topic and publishes cell counts (free, occupied, unknown), map dimensions, and resolution as \texttt{Float32MultiArray} messages. The Unity \texttt{SLAMTelemetry} component then derives coverage area and scan rate locally from these messages to update the dashboard.

\section{Experiments}
\label{sec:experiments}

\subsection{Setup}

Experiments were conducted in a 6$\times$8\,m laboratory. The Meta Quest~3 scanned the room with MRUK to generate a spatial mesh for passthrough occlusion. Three simulated TurtleBot3 robots were placed at distinct starting positions. The pipeline was evaluated across five 9-minute sessions, with a single operator teleoperating all three robots sequentially in each. The operator was a graduate researcher with prior ROS~2 and Unity experience but no prior exposure to the MR-SLAM interface. Table~\ref{tab:hardware} lists the configuration. An additional session in the same laboratory, recorded with the same hardware and software, provides the illustrative maps and timing distributions used in the figures of \S\ref{sec:results}.

\begin{table}[!htbp]
\centering
\caption{Hardware and Software Configuration}
\label{tab:hardware}
\small
\begin{tabular}{@{}ll@{}}
\toprule
\textbf{Component} & \textbf{Specification} \\
\midrule
MR Headset & Meta Quest~3 (Snapdragon XR2 Gen~2) \\
Unity Version & 2022.3 LTS (Android) \\
ROS~2 & Humble Hawksbill (Ubuntu 22.04) \\
SLAM & SLAM Toolbox~\cite{macenski_slam_2021} (async) $\times$3 \\
Map Merge & \texttt{multirobot\_map\_merge} (known poses) \\
Robot Model & TurtleBot3 Burger $\times$3 (simulated) \\
LiDAR & 180 rays, 90$^\circ$ FoV, 10\,Hz \\
Bridge & \texttt{ros\_tcp\_connector} (TCP:10000) \\
\bottomrule
\end{tabular}
\end{table}

\subsection{Metrics}

Five metrics were recorded using rosbag2 and \texttt{slam\_stats\_publisher}, selected to assess spatial fidelity (coverage, overlap agreement), communication reliability (scan rate), and timing consistency (TF jitter). \textbf{Coverage}: fraction of grid cells classified as free or occupied; coverage area is the observed-cell count multiplied by the squared cell resolution. \textbf{Cross-instance occupancy consistency}: for each robot pair, the percentage of spatially overlapping cells with identical occupancy values. With known initial poses and a shared \texttt{map} frame, this characterizes whether the three independent SLAM Toolbox solutions agree in shared regions, not whether unknown-pose merging would succeed. \textbf{Scan rate}: arrival frequency of \texttt{LaserScan} messages at SLAM Toolbox. \textbf{TF jitter}: absolute deviation of each Unity-published transform's clock difference from its per-link mean (excluding SLAM-published \texttt{map}$\rightarrow$\texttt{odom} links). \textbf{Teleop profile}: linear and angular velocities issued by the operator.

\begin{figure}[!htbp]
\centering
\includegraphics[width=\columnwidth]{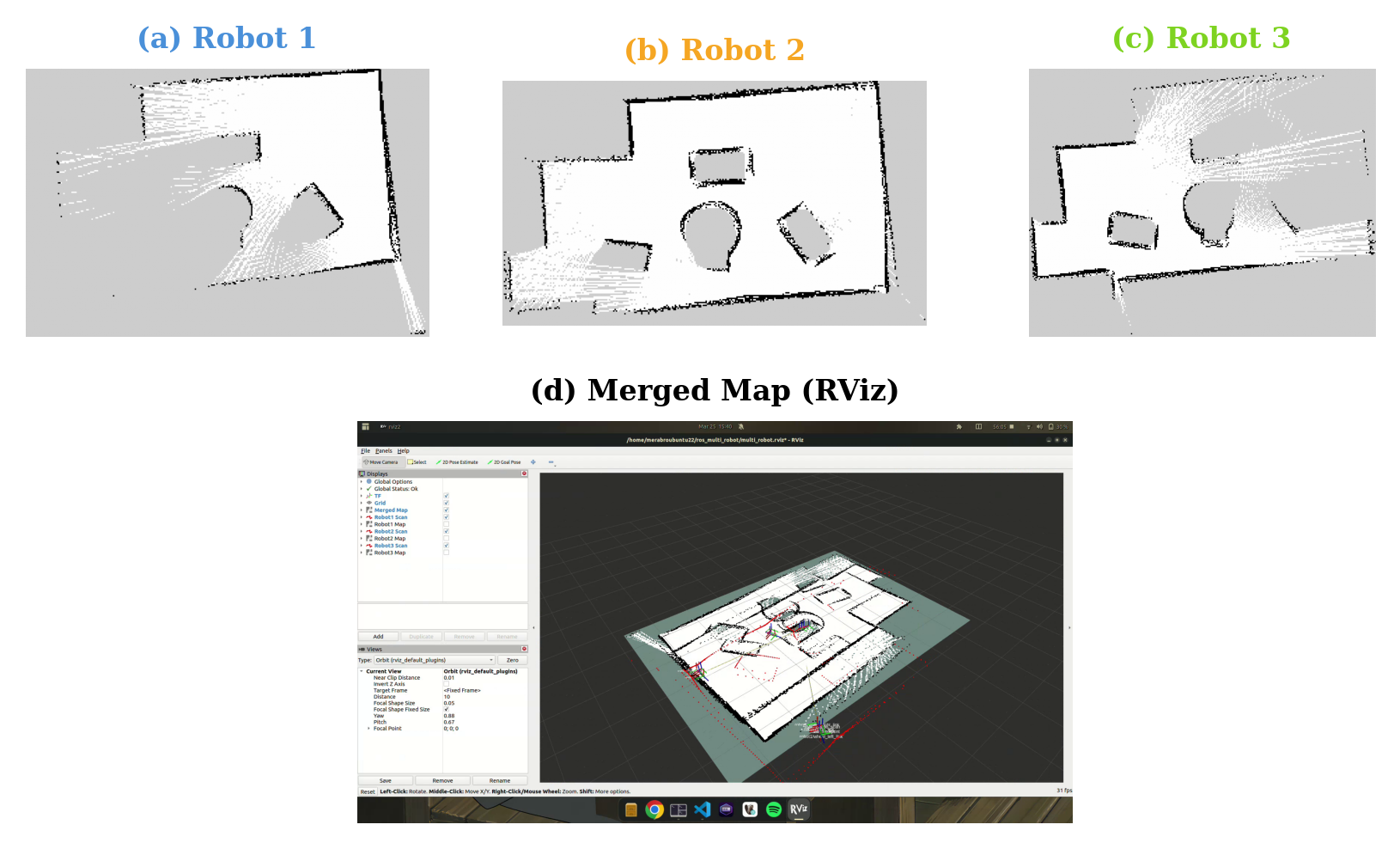}
\caption{Per-robot occupancy grids (top) and merged map from RViz (bottom). White: free, black: occupied, gray: unknown.}
\label{fig:maps}
\end{figure}

\subsection{Results}
\label{sec:results}

\begin{figure}[!htbp]
\centering
\includegraphics[width=\columnwidth]{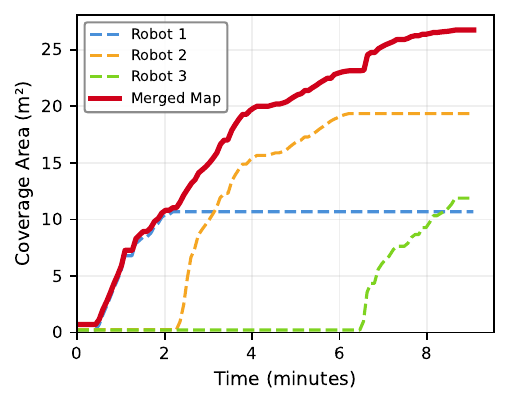}
\caption{Cumulative map coverage over time for each robot and the merged map.}
\label{fig:coverage}
\end{figure}

Across the five primary sessions, merged-map coverage averaged $17.9 \pm 0.8$\,m$^2$, with absolute coverage in m$^2$ depending on operator path and the specific room layout. Fig.~\ref{fig:coverage} shows cumulative coverage over time in the additional session, in which the merged map reached 26.7\,m$^2$ ($\sim$65\% of a 41\,m$^2$ grid). Robot~1 was driven first through the central area, followed by Robot~2 (left) and Robot~3 (right); coverage growth slowed as robots revisited previously mapped regions.

\begin{figure}[!htbp]
\centering
\includegraphics[width=0.49\columnwidth]{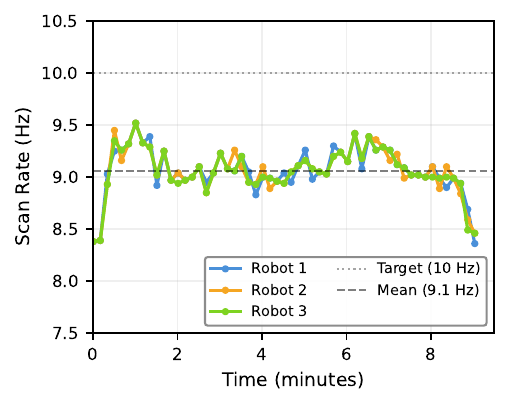}
\hfill
\includegraphics[width=0.49\columnwidth]{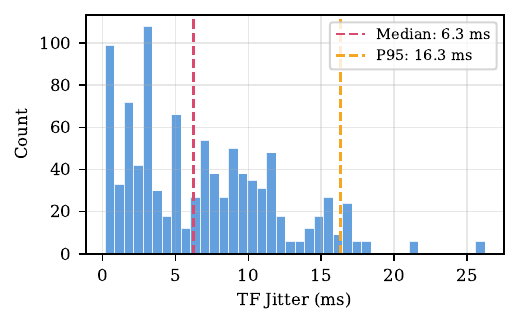}
\caption{(a)~Scan reception rates at SLAM Toolbox for each robot. (b)~TF jitter distribution; median 6.3\,ms, P95: 16.3\,ms.}
\label{fig:perf}
\end{figure}

Scan delivery averaged $8.83 \pm 0.16$\,Hz across the five primary sessions. Fig.~\ref{fig:perf}(a) shows the scan-rate trace from the additional session, with a mean of 9.06\,Hz and a minimum of 8.36\,Hz when the operator switched between robots, within 10\% of the commanded 10\,Hz. TF jitter (Fig.~\ref{fig:perf}b), measured in the additional session, had a median of 6.3\,ms (mean 6.8\,ms, P95: 16.3\,ms, max: 26.2\,ms) over 972 samples from 18 Unity-published links, consistent with the 82\,ms round-trip reported by Quest2ROS~\cite{welle_quest2ros_2024} and well below latency thresholds that cause SLAM divergence.

Cross-instance occupancy consistency between robot pairs averaged $94.7 \pm 0.5\%$ across the five primary sessions. In the additional session this value was 95.4\% (minimum 92.4\%) across 35 measurements where spatial overlap existed. Because the merge uses known initial poses and a shared \texttt{map} frame, this metric measures whether the three independent SLAM Toolbox instances produced coherent grids in shared regions, not the quality of the merge operation; it serves as a consistency check that no single instance has diverged. Fig.~\ref{fig:maps} shows the per-robot occupancy grids and the merged map from the additional session. Table~\ref{tab:results} summarizes the primary-session statistics.

\begin{table}[!htbp]
\centering
\caption{Quantitative Results across Five Sessions}
\label{tab:results}
\small
\begin{tabular}{@{}lc@{}}
\toprule
\textbf{Metric} & \textbf{Value (mean $\pm$ std, $n{=}5$)} \\
\midrule
Merged coverage & 17.9 $\pm$ 0.8\,m$^2$ \\
Cross-instance consistency & 94.7 $\pm$ 0.5\% \\
Scan rate & 8.83 $\pm$ 0.16\,Hz \\
TF jitter (additional session) & 6.3 / 16.3\,ms (median / P95) \\
Number of robots & 3 \\
Session duration & $\sim$9\,min each \\
\bottomrule
\end{tabular}
\end{table}

Fig.~\ref{fig:three_robots} shows the operator's passthrough view: virtual robots appear on the laboratory floor with correct perspective and occlusion. The dashboard panels (Fig.~\ref{fig:dashboard}) allow monitoring mapping progress without removing the headset.

\section{Discussion}
\label{sec:discussion}

\textbf{Centralized map merging.} Initial development attempted decentralized SLAM following Swarm-SLAM~\cite{lajoie_swarm-slam_2024}, but the geometrically clean raycasted scans lacked the feature distinctiveness needed for reliable cross-robot place recognition. Independent SLAM Toolbox~\cite{macenski_slam_2021} instances with centralized map merging proved more practical, at the cost of requiring known initial poses. This is acceptable in structured environments such as warehouses or offices; unstructured settings would require unknown-pose merging~\cite{carpin_fast_2008} or distributed optimization~\cite{tian_kimera-multi_2021}.

\textbf{Scalability and deployment.} The system supports three robots, limited by the operator's sequential teleoperation rather than computation (each SLAM instance is independent). Nav2 integration would decouple fleet size from operator bandwidth~\cite{gerkey_formal_2004, rizk_cooperative_2019}. Intended deployment scenarios are co-located fleet supervision tasks such as building-wing inspection and warehouse-aisle monitoring, where the operator is physically present in the workspace. We do not claim suitability for remote beyond-line-of-sight operation, where a non-passthrough rendering would be more appropriate. The effect of MR on operator cognitive load compared to conventional 2D displays is left to the planned NASA-TLX study (\S\ref{sec:conclusion}).

\textbf{Limitations.} The simulated LiDAR raycasts against Unity colliders; physical deployment would introduce sensor noise, odometry drift, and dynamic obstacles. MRUK's spatial occlusion uses an on-device mesh of the operator's room, reconstructed by the headset and independent of the robots' SLAM map. Headline numbers come from five 9-minute sessions with one operator in one laboratory; the figures use an additional session at the same location. Multi-operator and cross-environment evaluation remain future work.

\textbf{Spatial interaction.} The dashboard uses MR primarily through world-space anchoring of per-robot panels (Fig.~\ref{fig:dashboard}); the panel content itself is conventional 2D UI. Future MR-native extensions include gaze- or proximity-driven panel selection, in-situ rendering of the occupancy grid aligned to the operator's physical floor through passthrough, and hand-gesture waypoint placement in world space.

\section{Conclusion and Future Work}
\label{sec:conclusion}

This paper presented MR-SLAM, a mixed reality system for multi-robot SLAM in which an operator wearing a Meta Quest~3 teleoperates three simulated TurtleBot3 robots and monitors mapping progress through spatially anchored dashboards with passthrough occlusion. The system bridges Unity to ROS~2 Humble via \texttt{ros\_tcp\_connector}, running three independent SLAM Toolbox instances with real-time map merging. Across five 9-minute sessions, the system delivered scans at $8.83 \pm 0.16$\,Hz, mapped $17.9 \pm 0.8$\,m$^2$ of merged occupancy, and reached $94.7 \pm 0.5\%$ cross-instance occupancy consistency; an additional session contributed the 6.3\,ms median TF jitter measurement. The end-to-end pipeline (Algorithm~\ref{alg:tf}), including TF partitioning and QoS-aware clock synchronization, serves as a reusable reference for Unity--ROS~2 SLAM integrations.

Future work includes a multi-operator NASA-TLX study with a monitor-based baseline, unknown-pose map merging via spectral matching~\cite{carpin_fast_2008}, deployment on physical TurtleBot3 hardware, potential industrial validations in relevant industrial contexts \cite{KhatiriBWTP25}, and MR-native interaction features with Nav2 integration~\cite{gerkey_formal_2004, rizk_cooperative_2019}.

\balance
\bibliographystyle{IEEEtran}
\bibliography{references}

\end{document}